\documentclass{bmvc2k}

\title{RODEO: Replay for Online Object Detection}
\addauthor{Manoj Acharya}{ma7583@rit.edu}{1}
\addauthor{Tyler L. Hayes}{tlh6792@rit.edu}{1}
\addauthor{Christopher Kanan}{kanan@rit.edu}{1,2}

\addinstitution{Rochester Institute of Technology\\New York, USA}
\addinstitution{Paige\\New York, USA}

\runninghead{Acharya, Hayes, Kanan}{Replay for Online Object Detection}

\usepackage{graphicx}
\usepackage{comment}
\usepackage{amsmath,amssymb} %
\usepackage{color}
\usepackage{url}            %
\usepackage{booktabs}       %
\usepackage{amsfonts}       %
\usepackage{nicefrac}       %
\usepackage{microtype}      %

\usepackage{multirow}
\usepackage{blindtext}
\usepackage{placeins}
\usepackage{wrapfig,adjustbox}
\usepackage{xfrac}
\usepackage{enumitem}
\usepackage{caption}

\usepackage{array}
\usepackage{placeins}
\usepackage{rotating}
\usepackage{multicol}
\usepackage{multirow}
\usepackage{subfigure}
\usepackage{chngcntr}
\usepackage[linesnumbered,boxed,lined]{algorithm2e}
\usepackage{bm}

\usepackage{lipsum}
\usepackage{etoolbox}
\BeforeBeginEnvironment{wrapfigure}{\setlength{\intextsep}{0pt}}

\definecolor{Blue}{rgb}{.1,.1,.8}
\definecolor{Red}{rgb}{1,0,0}
\definecolor{Green}{rgb}{.1,.8,.1}

\usepackage{ifthen}
\newboolean{ack}
\newboolean{combined}
\setboolean{ack}{true} 
\setboolean{combined}{true}
\usepackage{pdfpages}

\newcommand{\beginsupplement}{%
        \setcounter{table}{0}
        \renewcommand{\thetable}{S\arabic{table}}%
        \setcounter{figure}{0}
        \renewcommand{\thefigure}{S\arabic{figure}}%
        
        \setcounter{section}{0}
        \renewcommand{\thesection}{S\arabic{section}}%
        
}

\begin{document}

\maketitle
\begin{abstract}
Humans can incrementally learn to do new visual detection tasks, which is a huge challenge for today's computer vision systems. Incrementally trained deep learning models lack backwards transfer to previously seen classes and suffer from a phenomenon known as ``catastrophic forgetting.'' In this paper, we pioneer online streaming learning for object detection, where an agent must learn examples one at a time with severe memory and computational constraints. In object detection, a system must output all bounding boxes for an image with the correct label. Unlike earlier work, the system described in this paper can learn  this task in an online manner with new classes being introduced over time. We achieve this capability by using  a novel memory replay mechanism that efficiently replays entire scenes. We achieve state-of-the-art results on both the PASCAL VOC 2007 and MS COCO datasets.

\end{abstract}

\section{Introduction}

Object detection is a localization task that involves predicting bounding boxes and class labels for all objects in a scene. Recently, many deep learning systems for detection~\cite{ren2015faster,redmon2017yolo9000} have achieved excellent performance on the commonly used Microsoft COCO~\cite{lin2014microsoft} and Pascal VOC~\cite{everingham2010pascal} datasets. These systems, however, are trained offline, meaning they cannot be continually updated with new object classes. In contrast, 
humans and mammals learn from non-stationary streams of samples, which are presented one at a time and they can immediately use new learning to better understand visual scenes. This setting is known as streaming learning, or online learning in a single pass through a dataset. Conventional models trained in this manner suffer from catastrophic forgetting of previous knowledge~\cite{mccloskey1989,french1999catastrophic}.

Streaming object detection enables new applications such as adding new classes, adapting detectors across seasons, and incorporating object appearance variations over time. Existing incremental object detection systems~\cite{Shmelkov_2017_ICCV,hao2019end,li2019rilod,shin2018incremental} have significant limitations and are not capable of streaming learning. Instead of updating immediately using the current scene, they update using large batches of scenes. These systems use distillation~\cite{hinton2015distilling} to mitigate forgetting. This means for the batch acquired at time $t$, they must generate predictions for all of the scenes in the batch before learning can occur, and afterwards they loop over the batch multiple times. This makes updating slow and impairs their ability to be used on embedded devices with limited compute or where fast learning is required. 

Previous works in incremental image recognition have shown that replay mechanisms are effective in alleviating catastrophic forgetting~\cite{hayes2019memory,rebuffi2016icarl,castro2018end,wu2019large}. Replay is inspired by how the human brain consolidates learned representations from the hippocampus to the neocortex, which helps in retaining knowledge over time~\cite{mcclelland1995there}. Furthermore, hippocampal indexing theory  postulates that the human brain uses an indexing mechanism to replay compressed representations from memory~\cite{teyler2007hippocampal}. In contrast, others replay raw samples~\cite{rebuffi2016icarl,castro2018end,wu2019large}, which is not biologically plausible. Here, we present the \textbf{R}eplay for the \textbf{O}nline \textbf{DE}tection of \textbf{O}bjects (RODEO) model, which replays compressed representations stored in a fixed capacity memory buffer to incrementally perform object detection in a streaming fashion. To the best of our knowledge, this is the first work to use replay for incremental object detection. We find that this method is computationally efficient and can be easily be extended to other applications. \textbf{This paper makes the following contributions:}
\begin{enumerate}[noitemsep,nolistsep]
    \item We pioneer streaming learning for object detection and establish strong baselines. 
    \item We propose RODEO, a model that uses replay to mitigate forgetting in the streaming setting and achieves better results than incremental batch object detection algorithms.
\end{enumerate}

\begin{figure}[t]
\begin{center}
\includegraphics[width=0.85\linewidth]{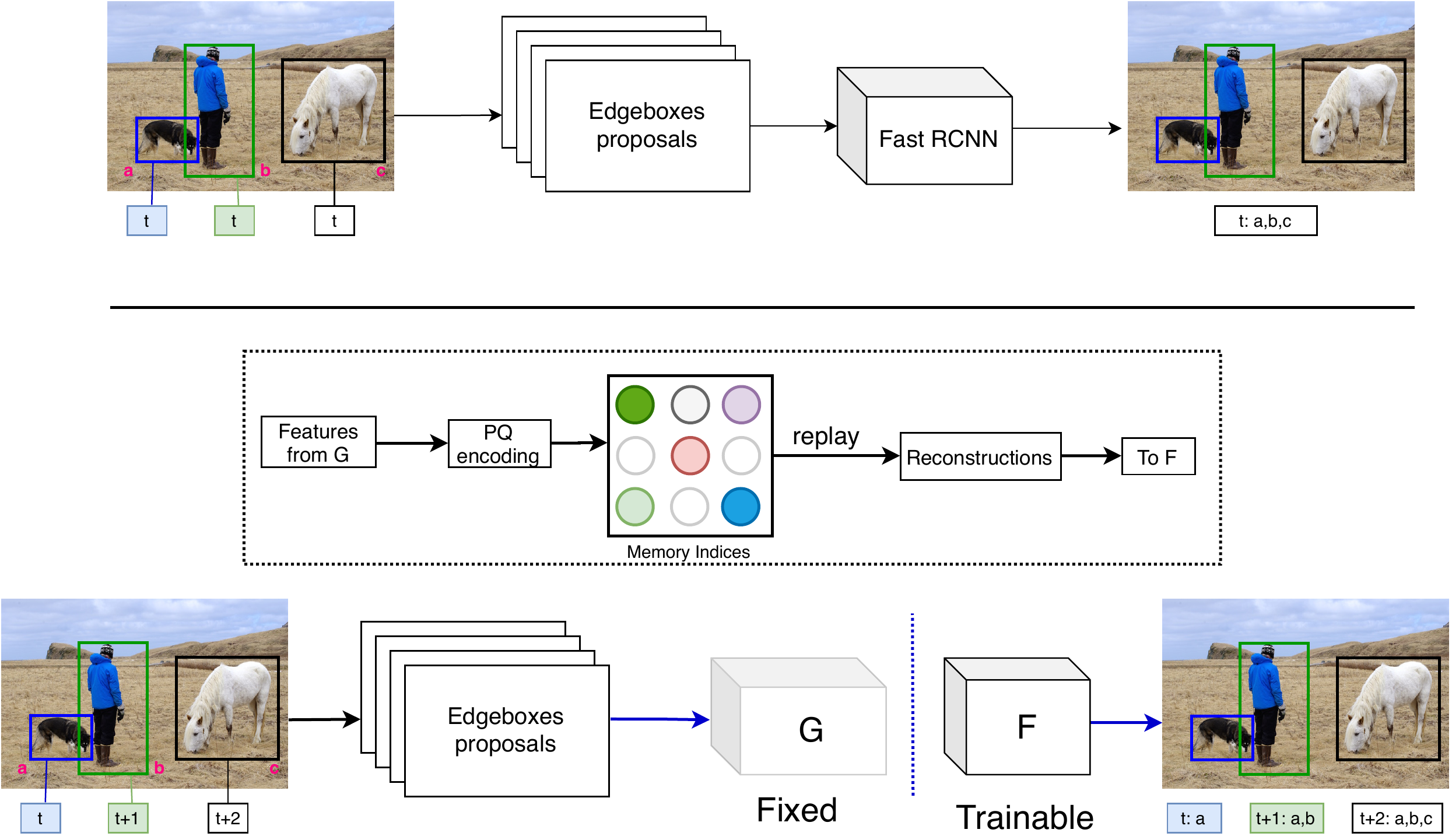}
\end{center}
\caption{In offline object detection, a model is provided an image and then trained with the ground truth boxes for all classes (e.g., a, b, c) in the image at once (top figure). However, in an online setting, ground truth boxes of different categories are observed at different time steps (bottom figure). While conventional models suffer from catastrophic forgetting, RODEO uses replay to efficiently train an incremental object detector for large-scale, many-class problems. Given an image, RODEO passes the image through the frozen layers of its network ($G$). The image is then quantized and a random subset of examples from the replay buffer are reconstructed. This mixture of examples is then used to update the plastic layers of the network ($F$) and finally the new example is added to the buffer.\label{fig:main}}
\end{figure}

\section{Problem Setup}

Continual learning (sometimes called incremental batch learning), is a much easier problem than streaming learning and has recently seen much success on classification and detection tasks~\cite{castro2018end,kemker2018fearnet,kirkpatrick2017,lopez2017gradient,nguyen2018variational,hou2019learning,wu2019large,chaudhry2018riemannian,lomonaco2019fine,parisi2019continual,Shmelkov_2017_ICCV}. In continual learning, an agent is required to learn from a dataset that is broken up into $T$ batches, i.e., $\mathcal{D}=\bigcup_{t=1}^{T}{B_t}$. At each time-step $t$, an agent learns from a batch consisting of $N_t$ training inputs, i.e., $B_t=\{I_i\}_{i=1}^{N_t}$ by looping through the batch until it has been learned, where $I_{i}$ is an image. Continual learning is not an ideal paradigm for agents that must operate in real-time for two reasons: 1) the agent must wait for a batch of data to accumulate before training can happen and 2) an agent can only be evaluated after it has finished looping through a batch. While streaming learning has recently been used for image classification~\cite{hayes2019lifelong,hayes2019memory,hayes2019remind,lopez2017gradient,chaudhry2019efficient}, it has not yet been explored for object detection, which we pioneer here.

More formally, during training, a streaming object detection model receives temporally ordered sequences of images with associated bounding boxes and labels from a dataset $\mathcal{D} = \{I_t\}_{t=1}^{T}$, where $I_{t}$ is an image at time $t$. During evaluation, the model must produce labelled bounding boxes for all objects in a given image, using the model built until time $t$. Streaming learning poses unique challenges for models by requiring the agent to learn one example at a time with only a single epoch through the entire dataset. In streaming learning, model evaluation can happen at any point during training. Further, developers should impose memory and time constraints on agents to make them more amenable to real-time learning.

\section{Related Work}

\subsection{Object Detection}
In comparison with image classification, which requires an agent to answer `what' is in an image, object detection additionally requires agents capable of localization, i.e., the requirement to answer `where' is the object located. Moreover, models must be capable of localizing multiple objects, often of varying categories within an image. Recently, two types of architectures have been proposed to tackle this problem: 1) single stage architectures (e.g., SSD~\cite{liu2016ssd,fu2017dssd}, YOLO~\cite{redmon2015you,redmon2017yolo9000}, RetinaNet~\cite{lin2017focal}) and 2) two stage architectures (e.g., Fast RCNN~\cite{wang2017fast}, Faster RCNN~\cite{ren2015faster}). Single stage architectures have a single, end-to-end network that generates proposal boxes and performs both class-aware bounding box regression and classification of those boxes in a single stage. While single stage architectures are faster to train, they often achieve lower performance than their two stage counterparts. These two stage architectures first use a region proposal network to generate class agnostic proposal boxes. In a second stage, these boxes are then classified and the bounding box coordinates are fine-tuned further via regression.
The outputs of all detection models are bounding box coordinates with their respective probability scores corresponding to the closest category. While incremental object detection has recently been explored in the continual learning paradigm, we pioneer streaming object detection, which is a more realistic setup.

\subsection{Incremental Object Recognition}

Although continual learning is an easier problem than streaming learning, both training paradigms suffer from catastrophic forgetting of previous knowledge when trained on changing, non-iid data distributions~\cite{mccloskey1989,french1999catastrophic}. Catastrophic forgetting is a result of the stability-plasticity dilemma, where an agent must update its weights to learn new information, but if the weights are updated too much, then it will forget prior knowledge~\cite{abraham2005memory}. There are several strategies for overcoming forgetting in neural networks including: 1) regularization approaches that place constraints on weight updates~\cite{hinton1987using,kirkpatrick2017,li2016learning,maltoni2018continuous,zenke2017continual,nguyen2018variational,chaudhry2018riemannian,aljundi2018memory}, 2) sparsity where a network sparsely updates weights to mitigate interference~\cite{FEL}, 3) ensembling multiple classifiers together~\cite{dai2007boosting,fernando2017,polikar2001learn,ren2017life,wang2003mining}, and 4) rehearsal/replay models that store a subset of previous training inputs (or generate previous inputs) to mix with new examples when updating the network~\cite{kemker2018fearnet,hayes2019memory,hayes2019remind,rebuffi2016icarl,castro2018end,hou2019learning,wu2019large}. Many prior works have also combined these techniques to mitigate forgetting, with a combination of distillation~\cite{hinton2015distilling} (a regularization approach) and replay yielding many state-of-the-art models for image recognition~\cite{rebuffi2016icarl,castro2018end,hou2019learning,wu2019large}.

\subsection{Incremental Object Detection}

While \emph{streaming} object detection has not been explored, there has been some work on object detection in the continual (batch) learning paradigm~\cite{Shmelkov_2017_ICCV,hao2019end,li2019rilod,shin2018incremental}. In \cite{Shmelkov_2017_ICCV}, a distillation-based approach was proposed without replay. A network would initially be trained on a subset of classes and then its weights would be frozen and directly copied to a new network with additional parameters for new classes. A standard cross-entropy loss was used with an additional distillation loss computed from the frozen network to restrict weights from changing too much. Hao et al.~\cite{hao2019end} train an incremental end-to-end variant of Faster RCNN~\cite{ren2015faster} with distillation, a feature preserving loss, and a nearest class prototype classifier to overcome the challenges of a fixed proposal generator. Similarly, \cite{li2019rilod} uses distillation on the classification predictions, bounding box coordinates, and network features to train an end-to-end incremental network. Shin et al.~\cite{shin2018incremental} introduce a novel incremental framework that combines active learning with semi-supervised learning. All of the aforementioned methods operate on batches and are not designed to learn one example at a time.

\section{Replay for the Online Detection of Objects (RODEO)}

Inspired by \cite{hayes2019remind}, RODEO is a model architecture that performs object detection in an online fashion, i.e., learning examples one at a time with a single pass through the dataset. This means our model updates as soon as a new instance is observed, which is more amenable to real-time applications than models operating in the incremental batch paradigm. To facilitate online learning, our model uses a memory buffer to store compressed representations of examples. These representations are obtained from an intermediate layer of the CNN backbone and compressed to reduce storage, i.e., compressed mid-network CNN tensors. During training, RODEO compresses a new image input. It then combines this new input with a random, reconstructed subset of samples from its replay buffer, before updating the model with this replay mini-batch.

More formally, our object detection model, $H$, can be decomposed as $H\left(\mathbf{x}\right) =  F\left(G\left(\mathbf{x}\right)\right)$ for an input image $\mathbf{x}$, where $G$ consists of earlier layers of a CNN and $F$ the remaining layers. We first initialize $G\left(\cdot\right)$ using a \emph{base initialization} phase where our model is first trained offline on half of the total classes in the dataset. After this base initialization phase, the layers in $G$ are frozen since earlier layers of CNNs learn general and transferable representations~\cite{yosinski2014transferable}. Then, during streaming learning, only $F$ is kept plastic and updated on new data.

\begin{algorithm}[ht]
 \caption{Incremental update procedure for RODEO on COCO.}
  \label{alg:rodeo-main}
\footnotesize
 \KwData{training set}
 \KwResult{train model parameters incrementally}
  Train entire object detection model offline on half of the dataset\;
  Train the PQ model on mid-CNN feature maps\;
  Initialize replay buffer with quantized samples from initialization\;
\For{$increment\leftarrow 41$ \KwTo $80$}{
 add new output units to classifier and box regressor\; 
\For{$image\leftarrow 1$ \KwTo $N$}{
  fetch edge box proposals\;
  fetch image annotation with ground truth boxes and labels\;
  push image through frozen layers and quantize\;
  \eIf{image in buffer}{
        append new image annotations to existing annotations\;
    }{add new quantized sample and annotations to buffer\;
        \If{buffer full}{
        remove an old sample and annotations from buffer\;
        }
    }
    reconstruct $n-1$ random samples from replay buffer\;
    train model on quantized current + replay ($n$) samples\;
    add current quantized sample to buffer\;
 }
 }
\end{algorithm}

Unlike previous methods for incremental image recognition~\cite{rebuffi2016icarl}, which store raw (pixel-level) samples in the replay buffer, we store compressed representations of feature map tensors. One advantage of storing compressed samples is a drastic reduction in memory requirements for storage. Specifically, for an input image $\mathbf{x}$, the output of $G \left(\mathbf{x} \right)$ is a feature map, $\mathbf{z}$, of size $p \times q \times d$, where $p \times q$ is the spatial grid size and $d$ is the feature dimension. After $G$ has been initialized on the base initialization set of data, we push all base initialization samples through $G$ to obtain these feature maps, which are used to train a product quantization (PQ) model~\cite{jegou2010product}. This PQ model encodes each feature map tensor as a $p \times q \times s$ array of integers, where $s$ is the number of indices needed for storage, i.e., the number of codebooks used by PQ. After we train the PQ model, we obtain the compressed representations of all base initialization samples and add the compressed samples to our memory replay buffer. We then stream new examples into our model $H$ one at a time. We compress the new sample using our PQ model, reconstruct a random subset of examples from the memory buffer, and update $F$ on this mixture for a single iteration. We subject our replay buffer to an upper bound in terms of memory. If the memory buffer is full, then the new compressed sample is added and we choose an existing example for removal, which we discuss next. Otherwise, we just add the new compressed sample directly. For all experiments, we store codebook indices using 8 bits or equivalently 1 byte, i.e., the size of each codebook is 256. We use 64 codebooks for COCO and 32 for VOC. For PQ computations, we use the publicly available Faiss library~\cite{FAISS}. A depiction of our overall training procedure is given in Alg.~\ref{alg:rodeo-main}.

For lifelong learning agents that are required to learn from possibly infinite data streams, it is not possible to store all previous examples in a memory replay buffer. Since the capacity of our memory buffer is fixed, it is essential to replace less useful examples over time. We use a replacement strategy that replaces the image having the least number of unique labels from the replay buffer. We also experiment with other replacement strategies in Sec.~\ref{section:replacement_strategies}.

\section{Experimental Setup}

\subsection{Datasets}

We use the Pascal VOC 2007~\cite{everingham2010pascal} and Microsoft COCO~\cite{lin2014microsoft} datasets. VOC contains 20 object classes with 5,000 combined training/validation images and 5,000 testing images. COCO contains 80 classes (including all VOC classes) with 80K training images and 40K validation images, which we use for testing. We use the entire validation set as our test set.

\subsection{Baseline Models}

We compare several baselines using the Fast RCNN architecture with edge box proposals and a ResNet-50~\cite{He_2016_CVPR} backbone, which is the setup used in \cite{Shmelkov_2017_ICCV}. These baselines include:
\begin{itemize}[noitemsep, nolistsep]
    \item \textbf{RODEO} -- RODEO operates as an incremental object detector by using replay mechanisms to mitigate forgetting. Our main variant replays 4 randomly selected samples from its buffer at each time step. We use 32 codebooks for VOC and 64 for COCO, each of size 256.
    \item \textbf{Fine-Tune (No Replay)} -- This is a standard object detection model without a replay buffer that is fine-tuned one example at a time with only a single epoch through the dataset. This model serves as a lower bound on performance and suffers from catastrophic forgetting of previous classes.
    \item \textbf{ILwFOD} -- The Incremental Learning without Forgetting Object Detection model~\cite{Shmelkov_2017_ICCV} uses a fixed proposal generator (e.g., edge boxes) with distillation to incrementally learn classes. It is the current state-of-the-art for incremental object detection.
    \item \textbf{SLDA + Stream-Regress} -- Deep streaming linear discriminant analysis was recently shown to work well in classifying deep network features on ImageNet~\cite{hayes2019lifelong}. Since SLDA is only used for classification, we combine it with a streaming regression model to regress for bounding box coordinates. To handle the background class with SLDA, we store a mean vector per class and a background mean vector per class, along with a universal covariance matrix. At test time, a label is assigned based on the closest Gaussian in feature space, defined by the class mean vectors and universal covariance matrix. More details for this model are provided in supplemental materials. 
    \item \textbf{Offline} -- This is a standard object detection network trained in the offline setting using mini-batches and multiple epochs through the dataset. This model serves as an upper bound for our experiments.
\end{itemize}
All models use the same network initialization procedure. Similarly, all models are optimized with stochastic gradient descent with momentum, except SLDA. We were not able to replicate the results for ILwFOD, so we use the numbers provided by the authors for VOC and do not include results for COCO since our setup differs. While RODEO, SLDA+Stream-Regress, and Fine-Tune are all streaming models trained one sample at a time with a single epoch through the dataset, ILwFOD is an incremental batch method that loops through batches of data many times making it less ideal for immediate learning.

\subsection{Metrics}

We introduce a new metric that captures a model's mean average precision (mAP) at a 0.5 IoU threshold over time. This metric extends the $\Omega_{all}$ metric from \cite{kemker2018forgetting,hayes2019memory} for object detection and normalizes an incremental learner's performance to an optimized offline baseline, i.e.,
$\Omega_{\mathrm{mAP}} = \frac{1}{T} \sum_{t=1}^T \frac{\alpha_{t}}{\alpha_{\mathrm{offline},t}} \enspace ,
$
where $\alpha_{t}$ is an incremental learner's mAP at time $t$, $\alpha_{\mathrm{offline}, t}$ is the offline learner's mAP at time $t$, and there are $T$ total testing events. We only evaluate performance on classes learned until time $t$. While $\Omega_{\mathrm{mAP}}$ is usually between 0 and 1, a value greater than 1 is possible if the incremental learner performed better than the offline baseline. This metric makes it easier to compare performance across datasets of varying difficulty.

\subsection{Training Protocol}

In our training paradigm, the model is first initialized with half the total classes and then it is required to learn the second half of the dataset one class at a time, which follows the setup in \cite{Shmelkov_2017_ICCV}. We organize the classes in alphabetical order for both PASCAL VOC 2007 and COCO. For example, on VOC, which contains 20 total classes, the network is first initialized with classes 1-10, and then the network learns class 11, then 12, then 13, etc. This paradigm closely matches how incremental class learning experiments have been performed for classification tasks~\cite{rebuffi2016icarl,hayes2019remind}. For all experiments, the network is incrementally trained on all images containing at least one instance for the new class. This means that images could potentially be repeated in previous or future increments. When training a new class, only the labels for the ground truth boxes containing that particular class are provided. 

For incremental batch models, after base initialization, models are provided a batch containing all data for a single class, which they are allowed to loop over. Streaming models operate on the same batches of data, but examples from within the batch are observed one at a time and can only be observed once, unless the data is cached in a memory buffer. For VOC, after each new class is learned, each model is evaluated on test data containing at least one box of any previously trained classes. For COCO, models are updated on batches containing a single class after base initialization, which is identical to the VOC paradigm. However, since COCO is much larger than VOC and evaluation takes much longer, we evaluate the model after  every 10 new classes of data have been trained.

\subsection{Implementation Details}

Following \cite{Shmelkov_2017_ICCV}, we use the Fast RCNN architecture~\cite{wang2017fast} with a ResNet-50~\cite{He_2016_CVPR} backbone and edge box object proposals~\cite{zitnick2014edge} for all models, unless otherwise noted. Edge boxes is an unsupervised method for producing class agnostic object proposals, which is useful in the streaming setting where we don't know what types of objects will appear in future time steps. Specifically, we compute 2,000 edge boxes for an image. Following \cite{ren2015faster}, we first resize images to 800 $\times$ 1000 pixels. To determine whether a box should be labelled as background or foreground, we compute overlap with ground truth boxes using an IoU threshold of 0.5.  Then, batches of 64 boxes are randomly selected per image, where each batch must have roughly 25\% positive boxes (IoU $>$ 0.5). During inference, 128 boxes are chosen as output after applying a per-category Non-Maximal Supression (NMS) threshold of 0.3 to eliminate overlapping boxes. More parameter settings are in supplemental materials.

For each input image to RODEO, layer $G$ produces feature map tensors of approximate size 25 $\times$ 30 $\times$ 2048. Images from the base initialization classes (1-10) for VOC and (1-40) for COCO are used to train the PQ model. For VOC, we are able to fit all the feature maps in memory to train the PQ model. For COCO, it is not possible to fit all the images in memory, so we sub-sample 30 random locations from the full feature map of each image to train the PQ.
The ResNet-50 backbone has four residual blocks. We quantize RODEO after the third residual block, i.e., $F$ consists of the last residual block, the Fast RCNN MLP head composed of two fully connected layers, and the linear classifier and regressor. To make experiments fair, we subject RODEO's replay buffer to an upper limit of 510 MB, which is the amount of memory required by ILwFOD. For VOC, this allows RODEO to store a representation of every sample in the training set. For COCO, this only allows us to store 17,668 compressed samples. To manage the buffer, we use a strategy that always replaces the image with the least number of unique objects.

\section{Experimental Results}

\begin{wraptable}[13]{r}{0.5\textwidth}
\vspace{-15pt}
\caption{$\Omega_{\mathrm{mAP}}$ results for VOC and COCO.}
\label{table:object-detection-results}
\begin{center}
\begin{tabular}{lcc}
\toprule
\textsc{Method} & \textsc{VOC} & \textsc{COCO} \\ 
\midrule
Fine-Tune & 0.385 & 0.220 \\
ILwFOD & 0.787 & - \\
SLDA+Regress & 0.696 & 0.655 \\
RODEO (recon, $n=4$) & 0.853 & \textbf{0.829} \\
RODEO (recon, $n=12$) & \textbf{0.906} & 0.760 \\
\midrule
RODEO (real, $n=4$) & 0.911 & 0.870 \\
RODEO (real, $n=12$) & 0.914 & 0.812\\
\midrule
Offline & 1.000 & 1.000 \\
\bottomrule
\end{tabular}
\end{center}

\end{wraptable}
Our main experimental results are in Table~\ref{table:object-detection-results} and learning curves are in Fig.~\ref{fig:inc-od-pascal} and Fig.~\ref{fig:inc-od-coco} for VOC and COCO, respectively. We include results for RODEO models that use both real and reconstructed features. Real features do not undergo reconstruction before being passed through plastic layers, $F$. To normalize $\Omega_{\mathrm{mAP}}$, we use offline models that achieve final mAP values of 0.715 and 0.42 on VOC and COCO, respectively. Additional results are in supplemental materials.

For VOC, RODEO beats all previous methods just by replaying only four samples. Our method is much less prone to forgetting than other models, which is demonstrated by its performance at the final time step in Fig.~\ref{fig:inc-od-pascal}. The SLDA+Regress model is surprisingly competitive on both datasets without the need to update its backbone. For COCO, RODEO is run with four replay samples and outperforms the baseline models by a large margin. Further, across various replay sizes and replacement strategies (Table~\ref{table:incremental-coco-main}), we find that real features yield better results compared to reconstructed features.

\begin{table}[t]
	\begin{minipage}{0.48\linewidth}
    \begin{center}
      \includegraphics[width=\linewidth]{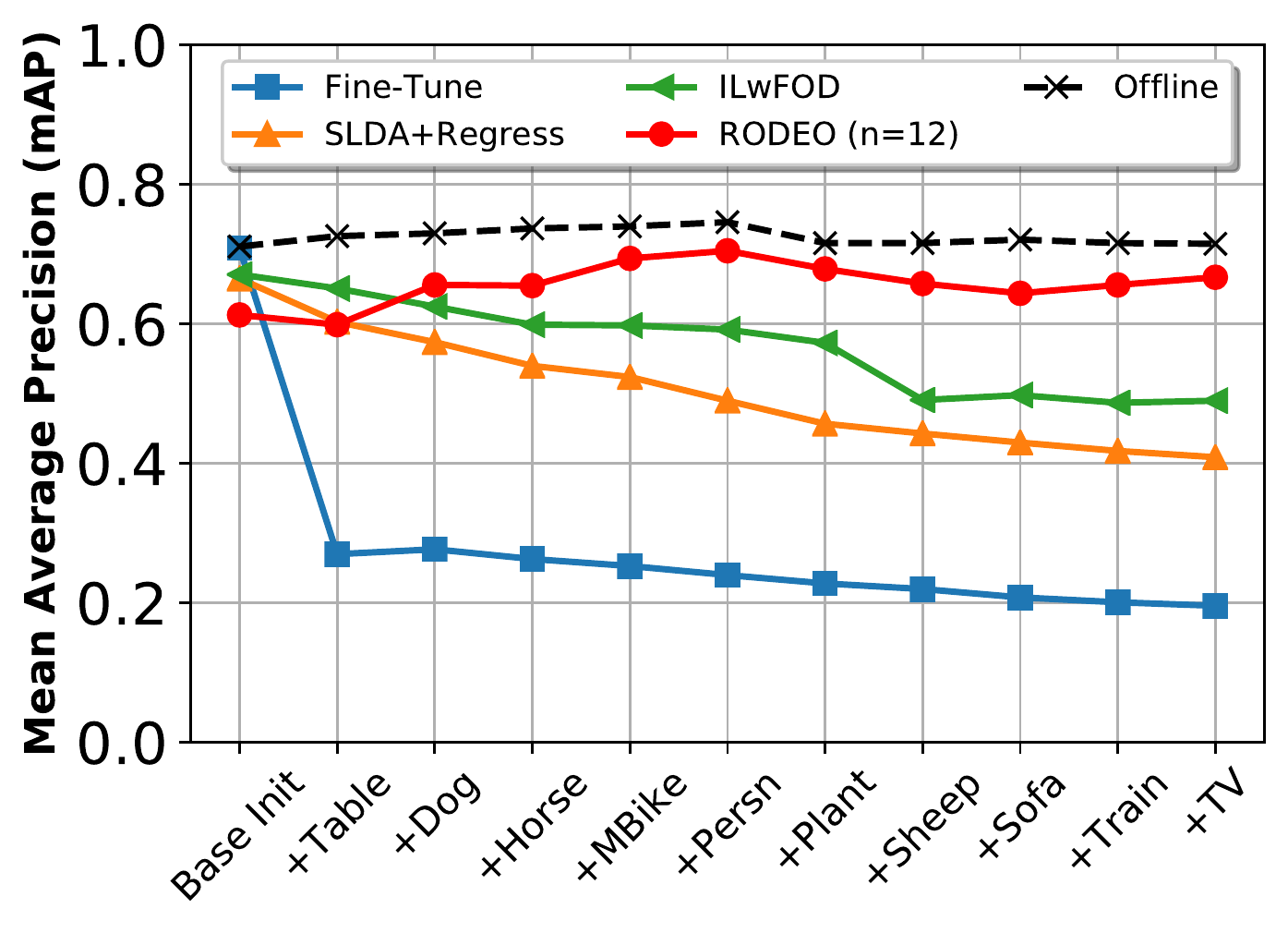}
    \end{center}
		\captionof{figure}{Learning curve for VOC 2007.}
		\label{fig:inc-od-pascal} 
	\end{minipage}\hfill
	\begin{minipage}{0.48\linewidth}
    \begin{center}
      \includegraphics[width=\linewidth]{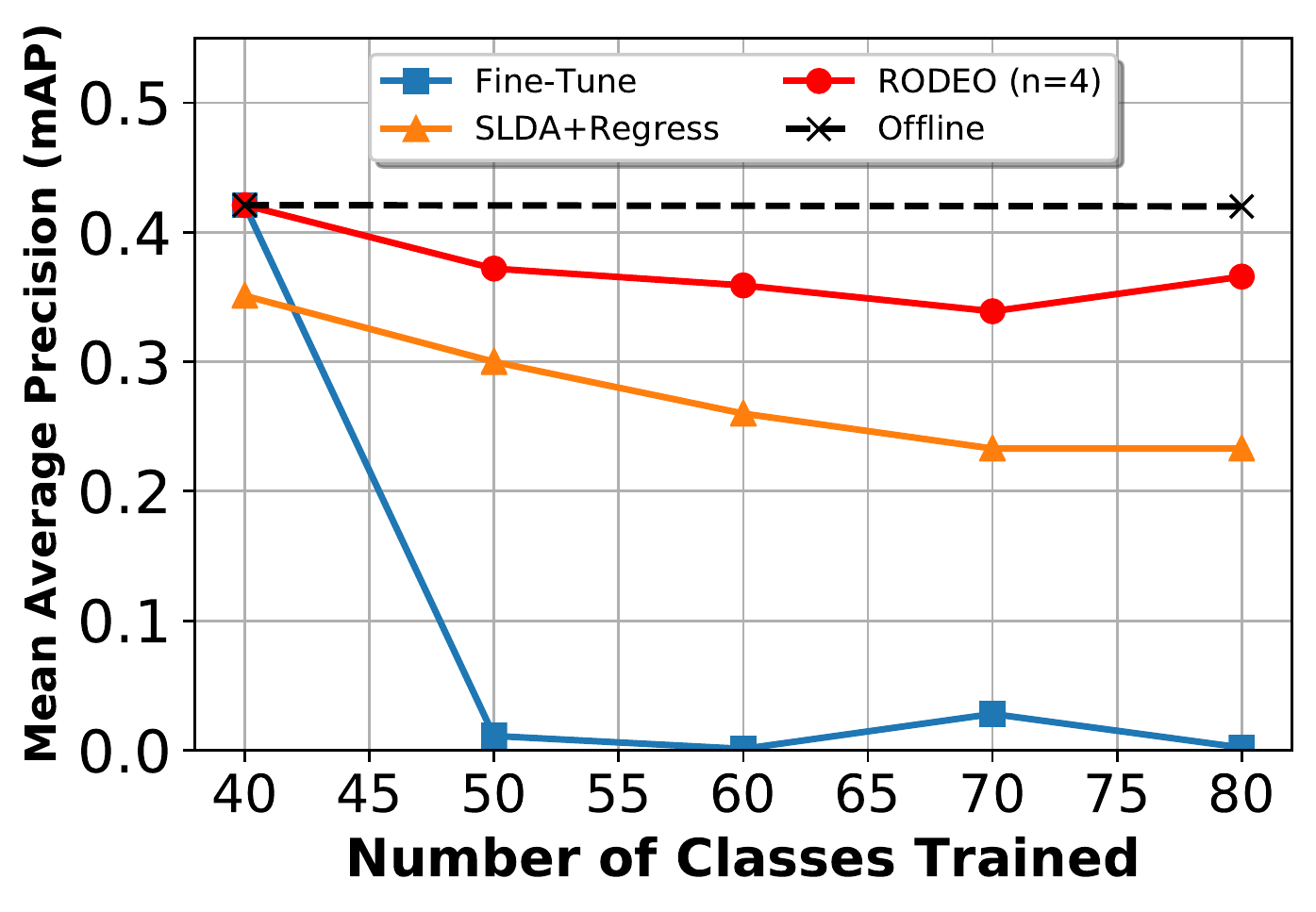}
    \end{center}
		\captionof{figure}{Learning curve for COCO.}
		\label{fig:inc-od-coco} 
	\end{minipage}
\end{table}

\begin{wraptable}[17]{r}{0.55\textwidth}
\caption{Incremental mAP results for several variants of RODEO.}
\label{table:incremental-coco-main}
\begin{center}
\footnotesize
\begin{tabular}{lcc}
\toprule
\textsc{Method} & \textsc{Mean} & \textsc{$\Omega_{\mathrm{mAP}}$} \\
\midrule
Fine-Tune &  0.093 & 0.220 \\
SLDA+Regress &  0.275 & 0.655 \\
RODEO $n=4$  (recon, BAL)  & 0.330 & 0.784  \\
\textbf{RODEO $n=4$ (recon, MIN)}  & \textbf{0.348} & \textbf{0.829} \\
RODEO $n=12$ (recon, BAL) & 0.312 & 0.741 \\
RODEO $n=12$ (recon, MIN)  & 0.320 & 0.760 \\
RODEO $n=4$ (recon, MAX)  & 0.119 & 0.282 \\
RODEO $n=4$ (recon, RANDOM) & 0.251 & 0.598 \\
\midrule
RODEO $n=4$  (real, BAL)  & 0.350 & 0.831 \\
\textbf{RODEO  $n=4$ (real, MIN)} & \textbf{0.366} & \textbf{0.870} \\
RODEO $n=12$ (real, BAL)  & 0.325 & 0.774 \\
RODEO  $n=12$ (real, MIN) & 0.342 & 0.812 \\
\midrule
RODEO  $n = 4$ (real, NO-REPLACE)  & 0.390 & 0.928 \\
\midrule
Offline & - & 1.000\\
\bottomrule
\end{tabular}
\end{center}
\end{wraptable}

\subsection{Additional Studies of RODEO Components}
\label{section:replacement_strategies}

To study the impact of the buffer management strategy chosen, we run the following replacement strategies on the COCO dataset. Results are in Table~\ref{table:incremental-coco-main}.
\begin{itemize}[noitemsep, nolistsep]
    \item \textbf{BAL:} Balanced replacement strategy that replaces the item which least affects the overall class distribution.
    \item  \textbf{MIN}, \textbf{MAX}: Replace the image having the least and highest number of unique labels respectively.
    \item \textbf{RANDOM:} Randomly replace an image from the buffer.
    \item \textbf{NO-REPLACE:} No replacement, i.e., store everything and let the buffer expand infinitely.
\end{itemize}

For an ideal case, we ran a version of RODEO with real features ($n=4$) and an unlimited buffer (storing everything). This model achieved an $\Omega_{\mathrm{mAP}}$ of 0.928. All other replacement strategies are only allowed to store 17,668 samples. We find that MAX replace yields even worse results compared to RANDOM replace suggesting storing more samples with more unique categories is better. Similarly, we find that MIN replace performs better across both real and reconstructed features, even beating the balanced (BAL) replacement strategy. We hypothesize that since MIN replace keeps images with the most unique objects, it results in a more diverse buffer to overcome forgetting.

For our VOC experiments, we do not replace anything from the buffer. As we increase the number of replay samples from 4 to 12, the performance improves by 0.3\% for real features and 5.3\% for reconstructed features respectively. Surprisingly for COCO, which has buffer replacement, the performance decreases as we increase the number of replay samples. We suspect this could be because COCO has many more objects per image compared to VOC which are being treated as background for region proposal selection. In the future, it would be interesting to develop new methods to handle this background class in an incremental setting, which has been explored for incremental semantic segmentation~\cite{cermelli2020modeling}.

\subsection{Training Time}
For COCO, we train each incremental iteration of Fast R-CNN for 10 epochs  which takes about 21.83 hours. Thus, full offline training of 40 iterations takes a total of 873 hrs. In contrast, our method, RODEO, requires only 22 hours which is a 40$\times$ speed-up compared to offline. SLDA+Regress and Fine-Tune both train faster, but perform much worse in terms of detection performance. These numbers do not include the base initialization time, which is the same for all methods. Exact numbers are in supplemental materials (Table~\ref{table:runtime-comparison}).

\section{Discussion}

In current object detection problem formulations, detected objects are not aware of each other. However, many real-world applications require an understanding of attributes and the relationships between objects. For example, Visual Query Detection (VQD) is a new visual grounding task for localizing multiple objects in an image that satisfies a given language query~\cite{acharya2019vqd}.  Our method can be easily extended for the VQD task by modifying the object detector to output only the boxes relevant to the language query.

In any real system where memory is limited, the choice of an ideal buffer replacement strategy is vital. For any agent that needs to learn new information over time, while also recalling previous knowledge, it is critical to store the most informative memories and replace those which carry less information. This procedure has also been studied in the reinforcement learning literature as experience replay~\cite{lin1992self,isele2018selective}. Our buffer size is limited because it is calculated with respect to the maximum storage required by the ILwFOD model~\cite{Shmelkov_2017_ICCV}. To efficiently use this limited storage, we tried various replacement strategies to store the newer examples such as: random replacement, class distribution balancing, and replacement of images with the most or fewest number of unique bounding boxes present. In the future, more efficient strategies for determining the maximum buffer size and replacement strategy could be useful for online applications.

RODEO is designed explicitly for streaming applications where real-time inference and overall compute are critical factors, such as robotic or embedded devices. Although RODEO uses Fast-RCNN, a two stage detector, which is slower than single stage detectors like SSD~\cite{liu2016ssd,fu2017dssd} and YOLO~\cite{redmon2015you,redmon2017yolo9000}, single stage approaches could be used to facilitate faster learning and inference. Moreover, RODEO currently uses a ResNet-50 backbone and can only process two images in a single batch. Using a more efficient backbone model like a MobileNet~\cite{sandler2018mobilenetv2} or ShuffleNet~\cite{ma2018shufflenet} architecture would allow the model to run faster with fewer storage requirements. In future work, it would be interesting to study how RODEO could be extended to single-stage detectors by replaying intermediate features and directly using the generated anchors instead of edge box proposals.

Further performance gains could be achieved by using augmentation strategies on the mid-level CNN features. Recently, several augmentation strategies have been designed explicitly for object detection~\cite{zoph2019learning,wang2019data,kisantal2019augmentation} and it would be interesting to explore how they could improve performance within deep feature space for an incremental learning application.

\section{Conclusion}

We proposed RODEO, a new method that pioneers streaming object detection. RODEO uses replay of quantized, mid-level CNN features to mitigate catastrophic forgetting on a fixed memory budget. Using our new model, we achieve state-of-the-art performance for incremental object detection tasks on the PASCAL VOC 2007 and MS COCO datasets when compared against models that operate in the easier incremental batch learning paradigm. Furthermore, our model is general enough to be applied to multi-modal incremental detection tasks in the future like VQD~\cite{acharya2019vqd}, which require an agent to understand scenes and the relationships between objects within them.

\ifthenelse{\boolean{ack}}{
\section*{Acknowledgements}
This work was supported in part by DARPA/MTO Lifelong Learning Machines program [W911NF-18-2-0263], AFOSR grant [FA9550-18-1-0121], and NSF award \#1909696. The views and conclusions contained herein are those of the authors and should not be interpreted as representing the official policies or endorsements of any sponsor.
}

\bibliography{egbib}

\ifthenelse{\boolean{combined}}{
\clearpage
\begin{center}
    {\Large Supplemental Material \normalsize}
\end{center}
\beginsupplement

\section{Training Details}

Hyper-parameter settings for RODEO and the offline models for VOC and COCO are given in Table~\ref{table:parameters}. Similarly, run time comparisons for the COCO dataset are in Table~\ref{table:runtime-comparison}.

\begin{table*}[h]
\caption{Training parameter settings for RODEO and offline models.}
\label{table:parameters}
\begin{center}
\footnotesize
\begin{tabular}{lccccc}
\toprule
\textsc{Parameters} & \textsc{VOC} & \textsc{COCO}\\
\midrule
Optimizer & SGD & SGD  \\
Learning Rate & 0.001 & 0.001 \\
Momentum & 0.9 & 0.9 \\
Weight Decay & 5e-4 & 5e-4 \\
Offline Batch Size & 2 & 2 \\
Offline Epochs & 25 & 10 \\
\bottomrule
\end{tabular}
\end{center}
\end{table*}

\begin{table*}[h]
\caption{Training-time comparison of models.}
\label{table:runtime-comparison}
\begin{center}
\footnotesize
\begin{tabular}{lc}
\toprule
\textsc{Method} & \textsc{Time(hour)} \\ 
\midrule
Fine-Tune & 4.2  \\
SLDA+Regress & 2.0 \\
RODEO & 21.7  \\
Offline & 873.2 \\
\bottomrule
\end{tabular}
\end{center}
\end{table*}

\section{Where to Quantize?}
Our choices of layers to quantize are limited due to the architecture of the ResNet-50 backbone. ResNet-50 has four main major layers with each having (3,4,6,3) bottleneck blocks respectively. Since bottleneck blocks add a residual shortcut connection at the end, it is not possible to quantize from the middle of the block, leaving only four places to perform quantization. Quantizing earlier has some advantages  since it leaves more trainable parameters for the incremental model, which could lead to better results~\cite{hayes2019remind}. But, it also requires twice the memory to store the same number of images as we move towards the earlier layers. For efficiency, we choose the last layer for feature quantization.

\section{Additional Results}

We provide the individual mAP results for each increment of COCO in Table~\ref{table:incremental-coco} and VOC in Table~\ref{table:incremental-voc}.

\begin{table*}[h]
\caption{Incremental mAP results for COCO evaluated after learning every 10 classes.}
\label{table:incremental-coco}
\begin{center}
\footnotesize
\begin{tabular}{lccccccc}
\toprule
\textsc{Method} & \textsc{1-40} & \textsc{50} & \textsc{60} & \textsc{70} & \textsc{80} & \textsc{Mean} & \textsc{$\Omega_{\mathrm{mAP}}$} \\
\midrule
Fine-Tune & 0.421 & 0.011 & 0.001 & 0.028 & 0.002 & 0.093 & 0.220 \\
SLDA+Regress & 0.351 & 0.300 & 0.260 & 0.233 & 0.233 & 0.275 & 0.655 \\
RODEO $n=4$  (recon,BAL) & 0.380  & 0.347 & 0.306 & 0.302 & 0.313 & 0.330 & 0.784  \\
\textbf{RODEO $n=4$ (recon, MIN)} & 0.380 & 0.355 & 0.353 & 0.325 & 0.329 & 0.348 & \textbf{0.829} \\
RODEO $n=12$ (recon, BAL) & 0.380 & 0.311 & 0.296 & 0.283 & 0.289 & 0.312 & 0.741 \\
RODEO $n=12$ (recon, MIN) & 0.380 & 0.317& 0.320 & 0.293 & 0.289 & 0.320 & 0.760 \\
RODEO $n=4$ (recon, MAX) & 0.380 & 0.015 & 0.060 & 0.064 & 0.074 & 0.119 & 0.282 \\ 
RODEO $n=4$ (recon, RANDOM) & 0.380 & 0.275 & 0.241 & 0.203 & 0.158 & 0.251 & 0.598 \\
\midrule
RODEO $n=4$  (real, BAL) & 0.421 & 0.356 & 0.333 & 0.313 & 0.326 & 0.350 & 0.831 \\
\textbf{RODEO  $n=4$ (real, MIN)} & 0.421 & 0.372 & 0.359 & 0.339 & 0.339 & 0.366 & \textbf{0.870} \\
RODEO $n=12$ (real, BAL) & 0.421 & 0.312 & 0.305 & 0.288 & 0.302 & 0.325 & 0.774 \\
RODEO  $n=12$ (real, MIN) & 0.421 & 0.328 & 0.330 & 0.318 & 0.312 & 0.342 & 0.812 \\
\midrule
RODEO  $n = 4$ (real, NO-REPLACE)  & 0.421 & 0.406 & 0.380 & 0.362 & 0.383 & 0.390 & 0.928 \\
\midrule
Offline & 0.421 & - & - & - & 0.420 & - & 1.000\\
\bottomrule
\end{tabular}
\end{center}
\end{table*}

\begin{table*}[h]
\caption{Incremental mAP results for the addition of each class in VOC dataset.}
\label{table:incremental-voc}
\begin{center}
\begin{adjustbox}{width=\linewidth,center}
\begin{tabular}{lccccccccccc}
\toprule
\textsc{Method} & \textsc{Base Init.} & \textsc{+table} & \textsc{+dog} & \textsc{+horse} & \textsc{+mbike} & \textsc{+persn} & \textsc{+plant} & \textsc{+sheep} & \textsc{+sofa} & \textsc{+train} & \textsc{+tv} \\
\midrule
Fine-Tune          & 0.709 & 0.270  & 0.277 & 0.263 & 0.253 & 0.24  & 0.228 & 0.220  & 0.208 & 0.201 & 0.196  \\
ILwFOD             & 0.671 & 0.651 & 0.625 & 0.599 & 0.598 & 0.592 & 0.573 & 0.491 & 0.498 & 0.487 & 0.490  \\
SLDA+Regress     & 0.665 & 0.603 & 0.574 & 0.540  & 0.524 & 0.490  & 0.457 & 0.443 & 0.430  & 0.418 & 0.409  \\
RODEO $n=4$ (recon)  & 0.614 & 0.596 & 0.582 & 0.602 & 0.650  & 0.667 & 0.635 & 0.581 & 0.635 & 0.620  & 0.617  \\
RODEO $n=12$ (recon) & 0.613 & 0.599 & 0.656 & 0.655 & 0.694 & 0.705 & 0.679 & 0.658 & 0.644 & 0.656 & 0.667  \\
\midrule
RODEO $n=12$ (real)  & 0.702 & 0.619 & 0.663 & 0.649 & 0.682 & 0.697 & 0.669 & 0.66  & 0.640  & 0.663 & 0.641 \\
RODEO $n=4$ (real)   & 0.702 & 0.619 & 0.629 & 0.665 & 0.678 & 0.701 & 0.671 & 0.649 & 0.640  & 0.661 & 0.646  \\
\midrule
Offline	        & 0.711	& 0.726 &	0.730 &	0.737	& 0.740 & 0.746 & 0.716 & 0.716 & 0.721 & 0.716 & 0.715 \\
\bottomrule
\end{tabular}
\end{adjustbox}
\end{center}
\end{table*}

\section{Additional SLDA+Stream-Regress Object Detection Details}

An overview of the incremental training stage for the SLDA+Stream-Regress object detection model is given in Alg.~\ref{alg:slda}. We use the Fast RCNN model to extract features from edge box proposals. Given a new input, we then make classification and regression predictions using the SLDA and Stream-Regress models, respectively. For both the SLDA model and the Stream-Regress models, we use shrinkage regularization with parameters of $1e-2$ and $1e-4$, respectively.

\begin{algorithm}[h]
 \KwData{training set}
 \KwResult{model fit to dataset}
 base initialization\;
 \For{image}{
  get edge box proposals\;
  get features, labels, and regression targets for edge box proposals\;
  get features and labels for ground truth\;
  freeze covariance matrix for SLDA model\;
  \For{box feat, label, regression targ in (edge box proposal features, edge box labels, edge box proposal regression targets)}{
  L2 normalize box feat\;
  \If{label is background}{
   fit SLDA model on box feat and specific background label\;
   }
  fit Stream-Regress model on box feat, label, and regression targ
  }
  unfreeze covariance matrix for SLDA model\;
  \For{box feat, label in (ground truth features, ground truth labels)}{
  L2 normalize box feat\;
  fit SLDA model on box feat and specific background label\;
 }
 }
 \caption{Incremental update procedure for SLDA+Stream-Regress.}
 \label{alg:slda}
\end{algorithm}

We train the SLDA model as proposed in \cite{hayes2019lifelong} with one slight modification. In \cite{hayes2019lifelong}, there was a single mean vector stored per class. However, in our work we allow SLDA to store two mean vectors per class, where one mean vector is representative of the actual class data and the second mean vector is representative of the background for that particular class. 
During test time, we thus obtain two scores for each class: the main class score and the background class score. We keep the main class score for each class and only keep the maximum score of all background scores.

Training the Stream-Regress model is similar to training the SLDA model. That is, we first initialize one mean vector $\bm{\mu}_{x} \in \mathbb{R}^{d}$ to zeros, where $d$ is the dimension of the data. We initialize another mean vector $\bm{\mu}_{y} \in \mathbb{R}^{m}$ to zeros, where $m$ is the number of regression targets, and we have four regression coordinates per class including the background class. We also initialize two covariance matrices, $\bm{\Sigma}_{x} \in \mathbb{R}^{d \times d}$ and $\bm{\Sigma}_{xy} \in \mathbb{R}^{d \times m}$, and a total count of the number of updates, $N \in \mathbb{R}$.

Given a new sample $\left(\mathbf{x}_{t}, \mathbf{y}_{t}\right)$, where $\mathbf{y}_{t} \in \mathbb{R}^{m}$ is a one-hot encoding of the regression targets, we make the following updates to our model:
\begin{equation}
        N = N + 1
\end{equation}    
\begin{equation}
    \mathbf{dx} = \mathbf{x}_{t} - \bm{\mu}_{x}
\end{equation}
\begin{equation}
    \mathbf{dy} = \mathbf{y}_{t} - \bm{\mu}_{y}
\end{equation}
\begin{equation}
    \bm{\Sigma}_{x} = \bm{\Sigma}_{x} + \frac{1}{N} \left(\frac{N-1}{N} \mathbf{dx}^{T} \mathbf{dx} - \bm{\Sigma}_{x}\right)
\end{equation}
\begin{equation}
    \bm{\Sigma}_{xy} = \bm{\Sigma}_{xy} + \frac{1}{N} \left(\frac{N-1}{N} \mathbf{dx}^{T} \mathbf{dy} - \bm{\Sigma}_{xy}\right)
\end{equation}
\begin{equation}
    \bm{\mu}_{x} = \bm{\mu}_{x} + \frac{\mathbf{dx}}{N}
\end{equation}
\begin{equation}
    \bm{\mu}_{y} = \bm{\mu}_{y} + \frac{\mathbf{dy}}{N} \enspace .
\end{equation}

To make predictions, we first compute the precision matrix 
\begin{equation}
    \bm{\Lambda} = \left[\left(1-\varepsilon\right)\bm{\Sigma}_{x} + \varepsilon \mathbf{I}\right]^{-1} \enspace ,
\end{equation}
with shrinkage parameter $\varepsilon$ and identity matrix $\mathbf{I} \in \mathbb{R}^{d \times d}$. We then compute regression targets, $\hat{\mathbf{r}} \in \mathbb{R}^{m}$, for an input $\mathbf{x}_{t}$ as:
\begin{equation}
    \hat{\mathbf{r}} = \mathbf{x}_{t}\mathbf{A} + \mathbf{b} \enspace ,
\end{equation}
where $\mathbf{A}=\bm{\Lambda}\bm{\Sigma}_{xy}$ and $\mathbf{b}=\bm{\mu}_{y}-\bm{\mu}_{x}\mathbf{A}$.

}

\end{document}